\documentclass[letterpaper, 10 pt, journal, twoside]{IEEEtran}

\IEEEoverridecommandlockouts                              

\usepackage{listings}
\usepackage{amsthm}

\usepackage{amsmath}
\usepackage{amssymb}
\usepackage{breqn}
\usepackage{mathtools,algorithm}
\usepackage[noend]{algpseudocode}
\usepackage{textcomp}
\usepackage{gensymb}
\usepackage{graphicx}
\usepackage{color}
\definecolor{ForestGreen}{RGB}{34,139,34}
\usepackage{varwidth}
\usepackage{microtype}
\usepackage{acronym}
\usepackage{hyperref}
\usepackage{multirow}
\usepackage{cite}
\usepackage{adjustbox}
\usepackage{glossaries}
\usepackage[noend]{algpseudocode}
\usepackage[symbol]{footmisc}
\usepackage{booktabs}
\usepackage[font=small,labelfont=bf,tableposition=top]{caption}
\usepackage{multicol}
\usepackage{blindtext}
\usepackage[normalem]{ulem}
\title{Robot Learning of Mobile Manipulation with \\ Reachability Behavior Priors}

\author{Snehal Jauhri, Jan Peters, and Georgia Chalvatzaki%
	\thanks{Manuscript received: February, 24, 2022; Revised May, 19, 2022; Accepted June, 17, 2022.}
	\thanks{This paper was recommended for publication by Editor Markus Vincze upon evaluation of the Associate Editor and Reviewers' comments.
		This work was supported by the German Research Foundation (DFG) Emmy Noether Programme (\#448644653) and the RoboTrust project of the Centre Responsible Digitality Hessen, Germany.}
	\thanks{All authors are with the Computer Science Department, Technische Universit\"{a}t Darmstadt, Germany {\tt\footnotesize \{snehal,jan,georgia\}@robot-learning.de}}%
	\thanks{Digital Object Identifier (DOI): 10.1109/LRA.2022.3188109.}%
	\thanks{© 2022 IEEE.  Personal use of this material is permitted.  Permission from IEEE must be obtained for all other uses, in any current or future media, including reprinting/republishing this material for advertising or promotional purposes, creating new collective works, for resale or redistribution to servers or lists, or reuse of any copyrighted component of this work in other works.}
}

\let\oldtwocolumn\twocolumn
\renewcommand\twocolumn[1][]{%
	\oldtwocolumn[{#1}{
		\begin{center}
			\includegraphics[width=\textwidth]{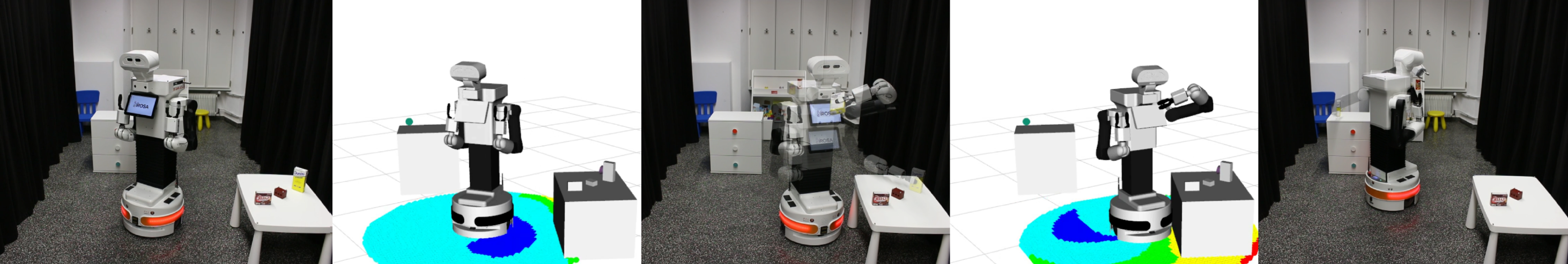}
			\captionof{figure}{Real-world execution of a 6D fetching and placing task with a TIAGo++. \textbf{Left:} Our robot has to choose a good base pose to pick up the yellow object. The visualization shows the learnt base Q-function of the robot. Dark {\color{blue}blue} signifies {\color{blue}maximum} likelihood of success while {\color{red}red} signifies the {\color{red}lowest}. \textbf{Middle:} The robot moves to the first sub-goal, picks up the object and queries the next base pose to place the object at the {\color{ForestGreen}green} location on top of the drawers. \textbf{Right:} The robot successfully places the object at the target location. }
			\label{fig:fetch_tiago}
		\end{center}
	}]
}

\begin{document}

	\newacronym{rl}{RL}{Reinforcement Learning}
\newacronym{mm}{MM}{Mobile Manipulation}
\newacronym{hyrl}{HyRL}{Hybrid RL}
\newacronym{bhyrl}{BHyRL}{Boosted HyRL}
\newacronym{irm}{IRM}{Inverse Reachability Map}
\newacronym{sac}{SAC}{Soft Actor-Critic}
\newacronym{kl}{KL}{Kullback-Leibler}
\newacronym{rpl}{RPL}{Residual Policy Learning}
\newacronym{mdp}{MDP}{Markov Decision Process}
\newacronym{lkf}{LKF}{learning kinematic feasibility}
\newacronym{ik}{IK}{Inverse Kinematics}
	\maketitle

	\begin{abstract}
		\gls{mm} systems are ideal candidates for taking up the role of personal assistants in unstructured real-world environments. \gls{mm} requires effective coordination of the robot's embodiments for tasks that require both mobility and manipulation. \gls{rl} holds the promise of endowing robots with adaptive behaviors, but most methods require large amounts of data. In this work, we study the integration of robotic reachability priors in actor-critic \gls{rl} methods for accelerating the learning of \gls{mm} for reaching and fetching tasks. Namely, we consider the problem of optimal base placement and the subsequent decision of whether to activate the arm for reaching a 6D target. We devise a novel \gls{hyrl} method that handles discrete and continuous actions jointly, resorting to the Gumbel-Softmax reparameterization. Next, we train a reachability prior using data from the operational robot workspace, inspired by classical methods. Subsequently, we derive \gls{bhyrl}, a novel actor-critic algorithm that benefits from  modeling Q-functions as a sum of residuals. For every new task, we transfer our learned residuals and learn the component of the Q-function that is task-specific, hence, maintaining the task structure from prior behaviors. Moreover, we find that regularizing the target policy with a prior policy yields more expressive behaviors. We evaluate our method in simulation in reaching and fetching tasks of increasing difficulty, and show the superior performance of \gls{bhyrl} against baseline methods. Finally, we zero-transfer our learned 6D fetching policy with \gls{bhyrl} to our \gls{mm} robot: TIAGo++. For more details, refer to our project site: \url{https://irosalab.com/rlmmbp}.
		
	\end{abstract}
	

	\section{Introduction}
	\IEEEPARstart{A}{utonomous} robots are expected to be a functional part of everyday living in the near future. Nevertheless, the ability of embodied agents to perform challenging tasks is very limited to static setups and repeated actions. \acrfull{mm} robots are an emblematic example of embodied AI systems that can incorporate the benefits of mobility and dexterity, thanks to their enlarged workspace and their equipment with various sensors. We envision such robots performing everyday living tasks like tidying up a room, setting up the dinner table etc.
	
	While significant research on \gls{mm} was delivered in the last decades \cite{brock_mobility_2016}, the limitation of the proposed algorithms to structured, well-defined environments does not allow the extrapolation of such methods to unstructured real-world setups. Recent breakthroughs in reinforcement and imitation learning \cite{haarnoja2018soft, ravichandar2020recent}, led to an increased deployment of learning methods in robotics \cite{ibarz2021train,lee_learning_2020,dalal_accelerating_2021}, with some early results on \gls{mm} tasks too \cite{li_hrl4in_2019,xia_relmogen_2021,honerkamp_learning_2021}. Among many challenges, one major issue in \gls{mm} is embodiment coordination, i.e., the coordinated motion of base, arms, torso, head, etc., for accomplishing a manipulation task. 
	
	This paper studies the integration of reachability priors in the process of learning coordinated \gls{mm} behaviors for reaching and fetching tasks. Notably, recent works in robot \gls{rl} explore the integration of behavior priors in the learning process intensely, with the purpose of providing algorithms that, on the one part, are sample-efficient, and on the other part, can be safer in terms of their exploration strategies when collecting experience for a new task \cite{nair2020awac,johannink2019residual,morgan2021model}. In the context of \gls{mm}, classical approaches would consider planning, and a task scheduler that coordinates the robot actions \cite{wolfe_combined_2010}, or an \gls{irm} that can indicate the areas where there is a higher chance for an \gls{mm} robot to reach a point, leading to some greedy trial-and-error of possible good base placements \cite{vahrenkamp_robot_2013,makhal_reuleaux_2018}. The learning-based methods employ deep \gls{rl} algorithms to learn in an end-to-end fashion \gls{mm} behaviors, however leading to impractically long training times \cite{honerkamp_learning_2021,xia_relmogen_2021}, that could be hazardous to a real robotic system. Real-world \gls{rl} training of \gls{mm} was only showcased in simplistic scenarios with a reduced action space and on a setup that is not easily transferable to robots with higher degrees of freedom and more complex structure \cite{sun2022fully}.
	
	In this work, we present a novel algorithm for robot learning of \gls{mm} using reachability priors, which inform about promising base locations that would lead to success in reaching or fetching tasks in 6D space. 
	We propose \gls{hyrl} for extending actor-critic methods to hybrid action spaces, to effectively have a single agent controlling both actions regarding the next pose and embodiment activation. We argue that the decision about embodiment activation is tightly connected to the robot pose. In this work, our action-space combines the decision over the next base pose with arm activation for reaching or grasping.
	
	Crucially, we study the integration of reachability priors for \gls{mm} that has limited access to future extended state spaces in new tasks, i.e., we study the integration of prior knowledge in an information asymmetric setting \cite{galashov2019information}. In essence, our prior knows only the relative 6D goal pose w.r.t. the robot and does not have any additional information in more complex settings where obstacles exist. To this end, we propose using residual Q-functions that effectively allow better transfer between complex settings; they can preserve underlying structure from previous tasks and be more robust to the information gap between the prior and the current policy. Additionally, we study the common treatment of action priors by considering a \gls{kl} divergence as a regularizer in \gls{rl} and show that when combined with Q-residuals in an actor-critic setting can yield better performance with more expressive behaviors, through our proposed algorithm: Boosted Hybrid Reinforcement Learning (BHyRL).
	
	Summarizing our contributions for robot learning of \gls{mm}:
	\begin{itemize}
		\item we propose to use a hybrid action-space reinforcement learning algorithm for effectively tackling the need for discrete and continuous action decisions in \gls{mm}
		\item we learn a reachability behavioral prior for mobile manipulation that can speed up the learning process, and incentivize the agent to select kinematically reachable poses when dealing with 6D reaching and fetching tasks, and
		\item we propose a new algorithm for transferring knowledge from behavior priors by modeling Q-functions as sums of residuals that \textit{boosts} the learning process, while also regularizing the policy learning in a trust-region fashion.
	\end{itemize}
	We evaluate our contributions in representative simulated tasks with the \gls{mm} robot TIAGo++ (Fig. \ref{fig:fetch_tiago}), and we provide extensive results and comparisons with baseline methods. Moreover, we show that our algorithmic contributions can scale to complex scenes with obstacles, while not forgetting previous behaviors, thanks to our boosted hybrid actor-critic \gls{rl} method. Due to our training process, we can transfer our learned behaviors in the real world for fetching objects with our TIAGo++ mobile manipulator robot \footnote{For more details and code release, please refer to our project site: \url{https://irosalab.com/rlmmbp}}. 
	
	\section{Related Work}
	\noindent\textbf{Mobile Manipulation.} The nominal chapter on \gls{mm} \cite{brock_mobility_2016} analyzes the ongoing challenges that hinder \gls{mm} systems due to the uncertainty in the world, the high dimensionality of \gls{mm} tasks (large workspace/configuration space), the need for discrete and continuous decisions (e.g., which embodiment to use), and generalization of \gls{mm} skills across tasks. Those challenges still persist, as identified by \cite{roa_mobile_2021}. While classical approaches have tried to leverage knowledge about the system to compute the reachability of \gls{mm} robots \cite{vahrenkamp_robot_2013, vahrenkamp_representing_2015, hertle_identifying_2017, welschehold_coupling_2018}, those do not consider the success of the task at hand and can only handle well-structured scenes. While task and motion planning can be coupled with \gls{mm} tasks, the acquisition of generalizable behaviors is still a challenge \cite{chitta_mobile_2012, wolfe_combined_2010, burget_bi_2016}.
	
	Robot learning promises to endow robots with skills acquired through experience that can allow them to adapt to dynamic environments reactively. Learning from human demonstrations can provide specific skills by imitation \cite{welschehold_learning_2017,welschehold_combined_2019}, however, those are limited to the provided data, and cannot extrapolate to new task instances. \gls{rl} utilizes exploration and learning by accounting for task success, and can therefore learn complex behaviors while being reactive. Recently, several \gls{rl} algorithms were proposed for solving \gls{mm} tasks as interactive navigation, where the hierarchical structure would be employed to decide possible sub-goals for the arm or the base in \cite{li_hrl4in_2019,xia_relmogen_2021}, that would be executed either through \gls{rl} policies or by motion planning respectively.
	On the other part, \cite{honerkamp_learning_2021} learns a policy that controls the base velocity using an augmented state-space while maintaining a reward function that accounts for kinematic feasibility of the end-effector pose. Learning whole-body control for \gls{mm} seems to benefit from structural information \cite{kindle_whole-body_2020, mittal2021articulated}.
	
	\noindent\textbf{Learning with behavior priors.} The use of behavior priors in \gls{rl} arises from the need for guided exploration towards sample-efficient learning, as well as for the long-wished generalization of skills. These behavior priors appear either as policy residuals \cite{silver_residual_2019, johannink2019residual} during optimization, or as a residual that robotic systems can adaptively deploy to ensure safety \cite{rana_bayesian_2021,rana_multiplicative_2020}. Utilizing planning as well as uncertainty about the observation space can also be used as additional information while training, leading to sample-efficient learning \cite{morgan2021model, lee_learning_2020}. A behavior prior can alter the behavior policy of \gls{rl} algorithms from a uniform policy, as in maximum entropy exploration \cite{haarnoja2018soft}, into a directed exploration that is restricted by the representation power of the prior through \gls{kl} regularization between the agent policy and the prior behavior policy \cite{peters2010relative,abdolmaleki2018maximum,cheng2019control}. Even when the behavior policy is learned through offline \gls{rl}, or by suboptimal experts, \gls{kl} regularization is employed during the online learning \cite{nair2020awac,jeong2020learning}. The benefit of exploiting a library of skills (learned/primitives) for accelerating learning was presented in \cite{pertsch_accelerating_2020, dalal_accelerating_2021}. Crucially, most approaches consider that behavior priors have complete access to the same state space as a task-specific agent; however, this might not be the case in realistic scenarios. This information asymmetry between prior and task-specific policy was studied in \cite{galashov2019information}, but for online behavior policy distillation. 
	
	\section{Preliminaries}
	\subsection{Problem Statement}
	Let us assume a robot with 10 degrees of freedom being able to move its base, torso, and arm. Given a 6D point in a free space that needs to be reached by the robot, our problem consists in finding the appropriate base placement in $\text{\textit{SE}}(2)$, that allows the robot arm to find a path towards the 6D point in $\text{\textit{SE}}(3)$. In the context of learning to discover such good poses, we want to use behavior priors that account for the robot structure and its workspace, i.e., accounting for the reachability and manipulability of the robotic arm given a static base pose. Moreover, we need to decide when this base position is optimal for triggering the arm to reach the 6D point. 
	
	In addition to the \gls{mm} problem, and in the context of learning with behavior priors, we identified the following problem in current methodologies; most of the works that account for behavior priors consider a fixed and known state-space \cite{dalal_accelerating_2021,rana_bayesian_2021}, which is not the case when the agent has to operate in unstructured environments. This problem can be associated with information asymmetry \cite{galashov2019information}, as our prior may not have full access to the information (full state-space) of the environment, as this is only revealed to the current agent policy. Therefore, our problem extends also to finding a solution for closing the information gap between \textit{partially informed} behavior priors for learning \gls{mm} and the task-specific policy, for promoting sample-efficient learning.

	\subsection{Reinforcement learning with action priors}
	Let us consider a \gls{mdp} described by the tuple $\{\mathcal{S}, \mathcal{A}, P, r, \gamma, P_0 \}$, where $\mathcal{S}$ and $\mathcal{A}$ are state and action spaces, $P: \mathcal{S} \times \mathcal{A} \times \mathcal{S} \to \mathbb{R}_+$  is the state-transition probability function describing the dynamics, $P_0 \to \mathbb{R}_+$ is the initial state distribution, $r:  \mathcal{S} \times \mathcal{A} \to \mathbb{R} $ is a reward function and $\gamma \in [0,1)$ is the discount factor. We define a policy $\pi \in \Pi: \mathcal{S} \times \mathcal{A} \to \mathbb{R}$ as the probability distribution of the event of executing an action $a$ in a state $s$. 
	A policy $\pi$ induces an action-value function corresponding to the expected discounted return collected by the agent when executing action $a$ in state $s$, and following the policy $\pi$ thereafter: 
	\begin{align}
		Q^\pi(s,a)& \triangleq \mathbb{E}_{\pi} \left[\sum_{k=0}^\infty \gamma^k r_{i+k+1} \middle\vert s_i = s, a_i = a \right],
	\end{align}
	where $r_{i+1}$ is the reward obtained after the $i$-th transition. Solving an \gls{mdp} consists of finding the optimal policy $\pi^*$, i.e. the one maximizing the expected discounted return. Given an action prior probability distribution $q(a|s)$, our learning objective becomes a relative-entropy policy optimization problem, using a \gls{kl} regularization between the agent policy and the prior policy
	\begin{align} \label{eq:obj_KL}
		\mathcal{J}(\pi) =& \mathbb{E}_{\pi} \left[\sum_{k=0}^\infty \gamma^k r_{i+k+1} \middle\vert s_i = s, a_i = a \right]  \nonumber \\
		&-\gamma^k \text{KL}\left[\pi(a_i|s_i)||q(a_i|s_i) \right],
	\end{align}
	\noindent where the \gls{kl} term acts as a regularizer. 
	\\
	\subsection{Inverse reachability maps}
	Every robotic system with a specific structure and degrees of freedom has specific capabilities in terms of reach. As described in \cite{vahrenkamp_robot_2013}, we can compute the operational workspace of a robot offline in order to query it for filtering out bad actions that are infeasible for the robot. We can compute the operational workspace, accounting for the probability of finding a configuration that leads to a successful reach in the 6D space, also taking into account joint limits or self-collisions, which effectively represent the robot's reachable workspace. In particular, for \gls{mm} robots, we can consider the floating base case, and we can compute the inverse mapping of the operational workspace to account for the potential robot base poses that allow the reaching of a 6D target point by the robot's end-effector. We can store these data and query them during online processing w.r.t. to a goal to be reached by the robot arm, filter the data w.r.t. to the floor plane to finally acquire the target-specific \gls{irm} in $\text{\textit{SE}}(2)$  of the \gls{mm} robot. We refer to \cite{vahrenkamp_robot_2013, makhal_reuleaux_2018} for a detailed explanation. 
	
	\section{Boosted Hybrid Reinforcement Learning}
	\subsection{Hybrid action-space \gls{rl}}\label{sec:hyrl}
	\gls{mm} tasks are excellent examples of control tasks where we need to take both discrete and continuous action decisions. For example, the continuous action variable may refer to a velocity command, while the discrete parameter decides which embodiment of the robot to use. This problem is usually handled in two different ways: i. in a hierarchical way \cite{li_hrl4in_2019};
	ii. by treating all control variables as continuous, needing thresholding of values outside the \gls{rl} agent policy (as in SGP-R \cite{xia_relmogen_2021}). The use of hybrid action spaces in robot control, in particular when strict hierarchies do not necessarily apply and when we need to optimize for discrete and continuous actions simultaneously, has shown to be beneficial, yet challenging \cite{neunert2020continuous}, as it requires the design of weights to be assigned to the discrete variables of the categorical distribution. We follow recent advances on the continuous relaxation of discrete random variables \cite{maddison2016concrete} and propose the use of the Gumbel-Softmax reparameterization for modeling the distribution of discrete actions, effectively proposing a \gls{hyrl} algorithm. 
	
	Let us define the hybrid action space $\mathcal{A} =  \mathcal{A}^c \times \mathcal{A}^d $, where the subscripts $c$ and $d$ denote the continuous and discrete subspace, respectively, with $\mathcal{A}^c \in \mathbb{R}^n$, for $n$-dimensional continuous actions, and $\mathcal{A}^d = \{a^1, a^2, \dots , a^m \}$ as a set of $m$ discrete actions. Differently from the assumption of \cite{neunert2020continuous}, we consider the actions dependent, as the discrete decision variable is dependent on the choice of the continuous variable. Our policy $\pi_{\theta}(a|s)$ with $a \in \mathcal{A}$ can be factored using the continuous and discrete policies, as  
	\begin{align}
		\pi_{\theta}(a|s) &= \pi^c_{\theta_1}(a^c|s) \pi^d_{\theta_2}(a^d|s, a^c) \nonumber \\
		&= \prod_{a_i^c\in \mathcal{A}^c}\left[\pi^c_{\theta_1}(a_i^c|s) \prod_{a_j^d\in \mathcal{A}^d} \pi^d_{\theta_2}(a_j^d|s, a_i^c) \right],
	\end{align}
	We represent the continuous policy as a Gaussian distribution with $\pi^c_{\theta_1}(a_i^c|s) = \mathcal{N}(\mu_{i,\theta_1}(s), \sigma_{i,\theta_1}(s))$, and our discrete policy $\pi^d_{\theta_2}(a_i^c|s)$ as a categorical distribution with class probability weights $\omega_1, \omega_2, ..., \omega_m$.
	
	Following \cite{maddison2016concrete, jang2016categorical}, any categorical distribution can be effectively reparameterized with a Gumbel-Softmax distribution, from which we can draw samples $z_j$ following: 
	\begin{equation}
		z_j = \frac{\text{exp}((\log(\omega_j) + g_j)/\tau)}{\sum^{m}_{l=1}\text{exp}((\log(\omega_l) + g_l)/\tau)} \text{ for } j=1, \dots, m,
	\end{equation}
	\noindent where $g_1, \dots, g_m$ are i.i.d. samples drawn from a Gumbel(0,1) distribution, and $\tau$ is the temperature of the softmax which provides a smooth distribution for $\tau >0$, and, therefore, we can compute gradients w.r.t. $\omega$.
	Given the discrete actions as a categorical distribution, we can replace them with the Gumbel-Softmax samples, and we can do backpropagation to update the parameters of the discrete policy. Therefore, we can use a neural network to learn the distribution of the discrete policy and condition on the continuous actions on which the discrete actions depend. Consequently, any actor-critic \gls{rl} algorithm can be employed, using the same Q-function to update the hybrid policy in \gls{hyrl}. 
	
	\subsection{Boosted \gls{rl} with priors} 
	
	Behavior priors can provide effective guiding signals for sample-efficient and even safer robot learning. However, the use of the KL constraint can shape the learning of a target task, but it does not address the challenges of effective information transfer from a prior task to a new task. For example, a robot that learns to reach exploiting its redundancy resolution can provide much information for a follow-up grasping task, but that is not effectively reflected by the KL constraint of (\ref{eq:obj_KL}). 
	
	We treat our prior as an initial task, to which we fit an action distribution and a related Q-function. We then formalize a transfer method for learning more complex tasks based on our reachability prior. Effectively, any new task can use knowledge from previous tasks (that may contain a subset of subtasks of the new task) as priors to guide exploration and speed-up learning. In a recent work of curriculum \gls{rl},  Klink et al. \cite{klink2022boosted} rely on the concept of boosting \cite{freund1995boosting,tosatto2017boosted}, to decompose complex tasks into a tailored sequence of subtasks as a curriculum of increasing difficulty, and they model the Q-function of the task at hand as the sum of residuals learned on the previous tasks of the curriculum. The authors show that this model leads to increased representation power of the function approximator in value-based \gls{rl}, and prove superior approximation error bounds for the estimate of the optimal action-value function w.r.t. using a single action-value approximator.
	
	In the context of behavior priors, we propose to learn residuals of the Q-function, leveraging the knowledge of the prior task for transferring and accelerating learning in the new task, since the residual needs to approximate only a part of the TD target, while the prior retains the structure of the previously learned task. Based on these advances, we introduce our method for \gls{bhyrl} with priors. In \gls{bhyrl}, we propose an alternative learning objective for actor-critic \gls{rl} methods that employ priors both for structuring the Q-function as a sum of residuals and also regularizing the expressivity of the new-task policy while
	handling the information asymmetry between the different tasks.  
	
	\noindent \textbf{Critic update:} To estimate the action-value function $Q^T(s_t,a_t)$ of task T, we use residuals $\rho$ that are approximated with neural networks. $Q^T$ can be estimated recursively as
	\begin{equation}\label{eq:residuals}
		\left.
		\begin{array}{l}
			Q_{\phi_0}^0=\rho^0_{\phi_0}\\
			Q_{\phi_1}^1= \rho^0 + \rho^1_{\phi_1}\\
			\dots \\
			Q_{\phi_T}^T= \rho^0 + \rho^1 + \dots + \rho^T_{\phi_T}\\
		\end{array}
		\right \} Q^T_{\phi_T} = \sum^{T-1}_{i=0}\rho^i + \rho^T_{\phi_T},
	\end{equation}
	\noindent where $\phi_T$ denotes the learnable parameters of the residual network $\rho^T_{\phi_T}$ of task $T$. Accordingly, the Q-function of a task T is obtained by minimizing the loss
	\begin{equation}\label{eq:Qobj}
		\mathcal{L}(\phi^{Q^T}) = \mathop{\mathbb{E}}_{\substack{s, a, s', r \\ \sim D^T}} \left[(Q^T_{\phi}(s,a) - y^T)^2  \right],
	\end{equation}
	\noindent where $y^T = r(s,a) + \gamma Q^T(s', \pi^T(\cdot|s'))$ is the TD-target for task T. \textit{Every residual} that was trained in a prior task may have \textit{information asymmetry} w.r.t. the current task, i.e., it only has access to the part of the state that is relevant, though we omitted this notation in \eqref{eq:Qobj} for simplicity. Only the task-specific trainable residual has access to the full state of the task. However, even if the state-space changes the hybrid action-space remains the same, as described in Sec. \ref{sec:hyrl}.
	
	\noindent \textbf{Actor update:} The hybrid policy is trained on the updated Q-function, maximizing the following objective 
	\begin{equation}
		\small
		\mathcal{L}(\theta^{\pi^{T}}) = \mathop{\mathbb{E}}_{\substack{s \sim D^T,\\ a\sim \pi^T_\theta(\cdot|s)}}\left[ Q^T(s,a)\right] - \alpha \text{KL}(\pi^{T-1}(\cdot|s^{T-1})||\pi^T_{\theta}(\cdot|s)),
	\end{equation}
	\normalsize
	\noindent where $\theta = \{\theta_1, \theta_2\}$ are the parameters of the continuous and discrete policy respectively. $\pi^{T-1}(\cdot|s^{T-1})$ refers to the prior policy distribution of task $T-1$, that may have access only to the relevant state information $s^{T-1}$ instead of the full state $s$ of task $T$ (whenever applicable).
	
	We found that applying forward \gls{kl} regularization is more beneficial for the policy fitting, as it incentivizes the agent to match the relevant part of the prior due to information asymmetry, but still allows the agent to extrapolate to the new task.
	Moreover, the target policy is trained over the Q-residuals that contain the structure of all previous tasks, overall leading to more expressive policies.

	\subsection{Algorithmic details of \gls{bhyrl} for \gls{mm}}
	For the \gls{mm} tasks in this paper we constrain the continuous action space for base placement to a fixed radius around the robot, and consider a discrete decision variable to control arm activation. We use as initial prior an agent trained offline using the \gls{irm} map of TIAGo++ as proposal action distribution, and thus acquire a prior policy over configurations with high probability of finding an IK solution for reaching a 6D point in the robot's workspace. While the learned Q-function of the reachability prior serves as our starting residual $\rho_0$, the acquired reachability policy regularizes only the next task. Every new task we transfer to, uses the previous policy for regularization, as this policy was trained over the Q-residuals, hence, incorporating the knowledge of all prior tasks. For the implementation of \gls{bhyrl}, we adopted the \gls{sac} algorithm to leverage stochastic policies, but we explore with an added Gaussian noise $\sim \mathcal{N}(0, 0.1)$ to each action.
	
	\section{Experimental results}
	\subsection{Experimental setup}
\begin{figure}[t!]
		\vspace{+0.25em}
		\centering
		\includegraphics[width=\columnwidth]{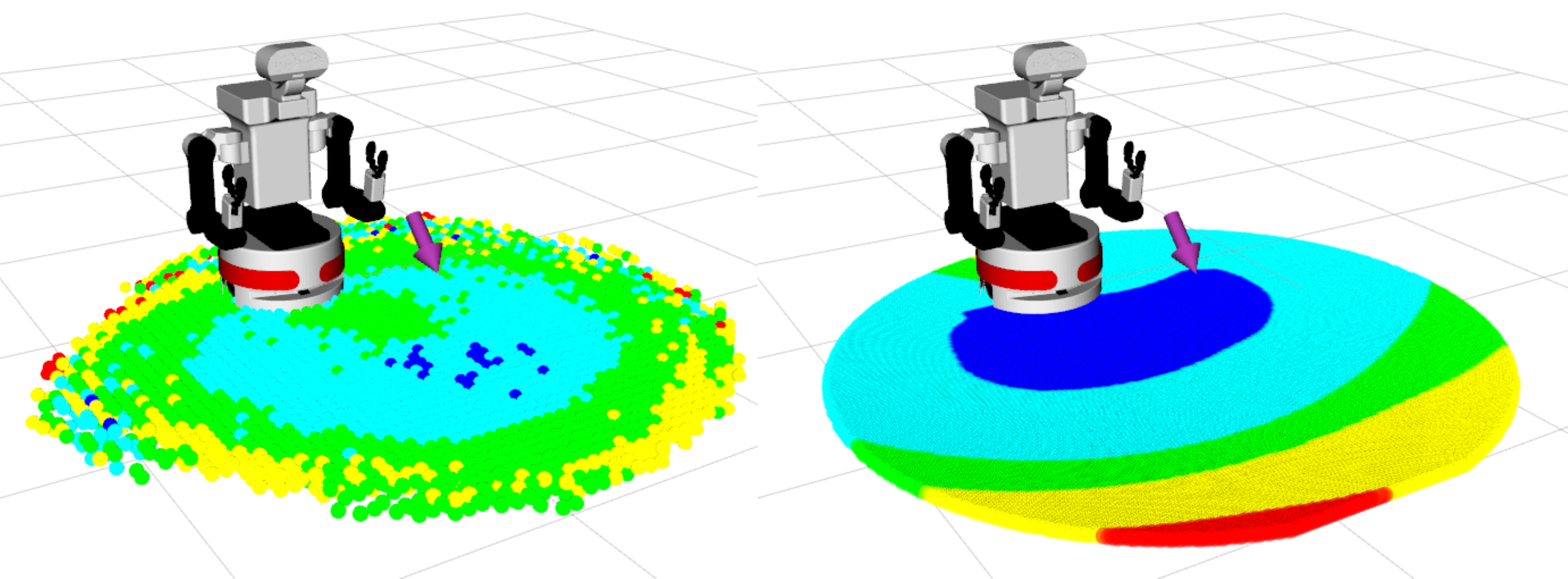}
		\captionof{figure}{Representation of reachability maps for the TIAGo++ robot, when considering a target pose (purple arrow) for the left arm. On the left, we visualize the \gls{irm} computed with manipulability measures as in \cite{vahrenkamp_robot_2013}. On the right, we depict our learned map (via Q-function querying) through \gls{hyrl}. Dark {\color{blue}blue} points are base locations with high reaching likelihood as per the maps.}
		\label{fig:IRMvsHyRL}
		\vspace{+0.35em}
		\vspace{+0.35em}
		\captionof{table}{Summary of hyperparameters for \gls{bhyrl}.}
		\adjustbox{max width=\textwidth}{
			\begin{tabular}{l|c}
				\toprule
				Hyperparameter   & Value  \\ \hline 
				\midrule
				discount $\gamma$ & $0.99$ \\ \hline
				Actor learning rate &  3e-4 \\ \hline
				Critic learning rate &  3e-4 \\ \hline
				[min, max] policy std &   [1e-3,1e3]        \\ \hline
				\gls{kl} weight $\alpha$ & 1e-3 \\ \hline
				Gumbel-Softmax $\tau$ & 1 \\ \hline
				$[w_1, w_2, w_3, w_4]$ &  [0.1,-0.05, -0.25,1] \\ \hline
				$[\textrm{IKpunish}, \textrm{Collision}, \textrm{task}]$ & [ 1, 1, 1] \\ \hline
				\bottomrule
		\end{tabular}}
		
		\label{tab:hyperparam}
		\vspace{-0.6cm}
	\end{figure}

	For the evaluation of \gls{bhyrl}, we created different reaching and fetching tasks in simulation. Our algorithmic implementation used the library MushroomRL \cite{MushroomJMLR}, while we developed our environments in \textit{Isaac Sim}. For studying the results of our proposed learning framework, we rely on the simulator state. We assume that we have access to a bounding box information about objects in the scene, and that we have a set of known grasps on the object.
	Though we simulate our bimanual \gls{mm} robot TIAGo++, in this work, we only consider \textit{the left arm}, and we will extend \gls{bhyrl} to using both arms in the future. For executing the arm actions, we compute the IK solutions considering \textit{left-arm-torso} kinematics.
	Table \ref{tab:hyperparam} enlists the hyperparameters of \gls{bhyrl}. Our reward functions can be described as 
	\begin{align}
		r(s,a) =& w_1  \textrm{deltaDist}(s,a) + w_2 \textrm{IKpunish}(s,a)  \nonumber \\
		& + w_3 \textrm{Collision}(s,a) +w_4 \text{task}(s,a),
	\end{align}
	
	\noindent where $w$ are weights scaling the rewards. $\textrm{deltaDist}(s,a)$ is the translational distance covered when taking action $a$ w.r.t. the goal. Note that the rotational distance is not included here. $\textrm{IKpunish}(s,a)$ adds a punishment for IK failures. This  motivates the agent to learn to use the arm i.e., query the IK, only when it is likely to succeed and avoid unnecessary IK queries, for eg., when the goal is beyond reach.
	$\textrm{Collision}(s,a)$ adds a punishment to collisions with any objects, and $\text{task}(s,a)$ rewards task success.
	
	We developed tasks of different difficulty, for evaluating the transferability of prior knowledge to new tasks. We provide extensive comparisons with baseline methods both in the context of learning \gls{mm}, and w.r.t. different learning algorithms for prior policy deployment. As different tasks and methods, require different reward functions, action spaces, etc., we rely on the metrics of \textit{success rate} and \textit{average action queries per episode} over $5$ seeds for evaluation. Finally, we test the transferability of the method to a real-world application of \gls{mm} for object fetching with TIAGo++.
	
	\begin{figure*}[t!]
		\vspace{+0.5em}
		\begin{tabular}{ c c c }
			\includegraphics[width=.30\textwidth]{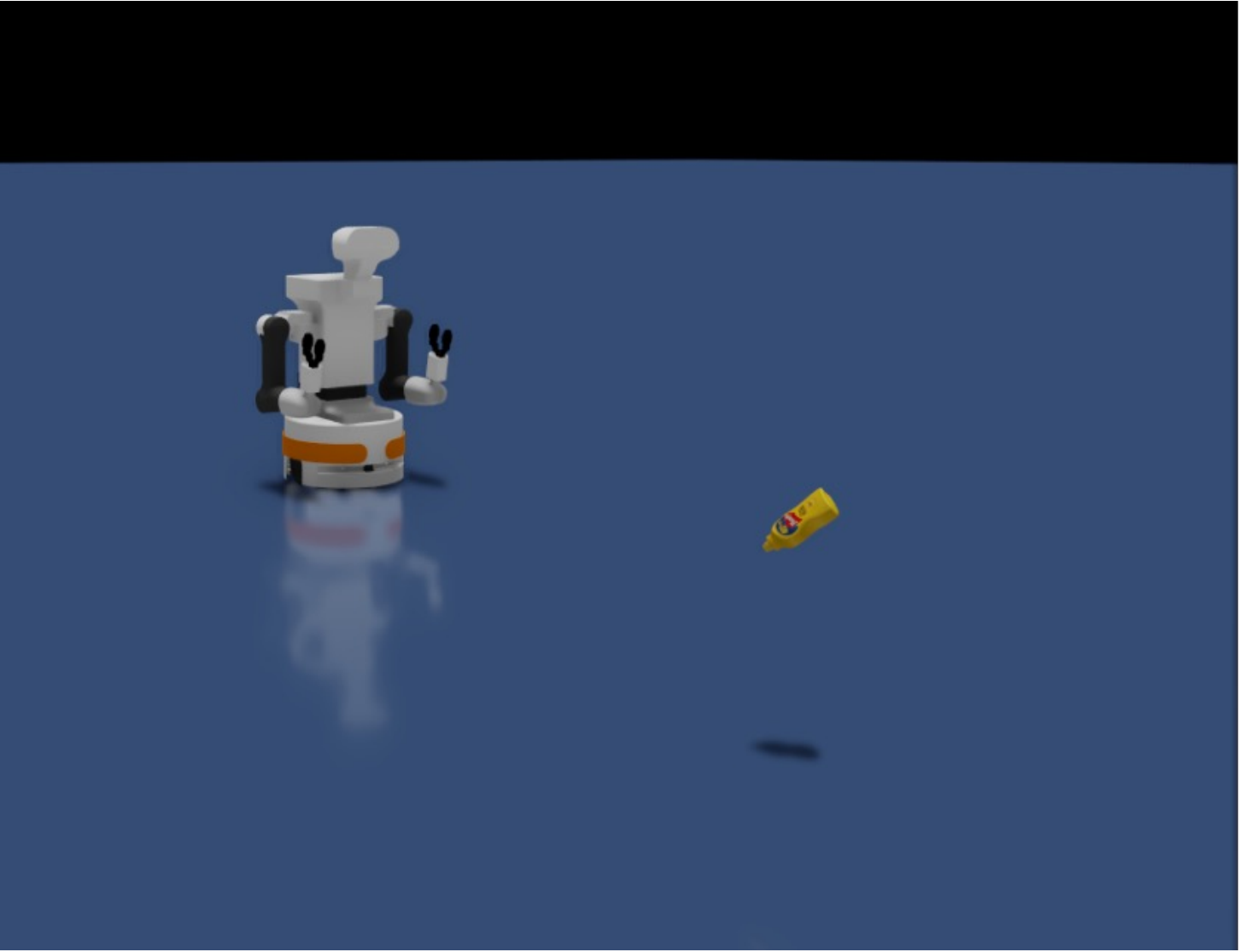} &
			\includegraphics[width=.31\textwidth]{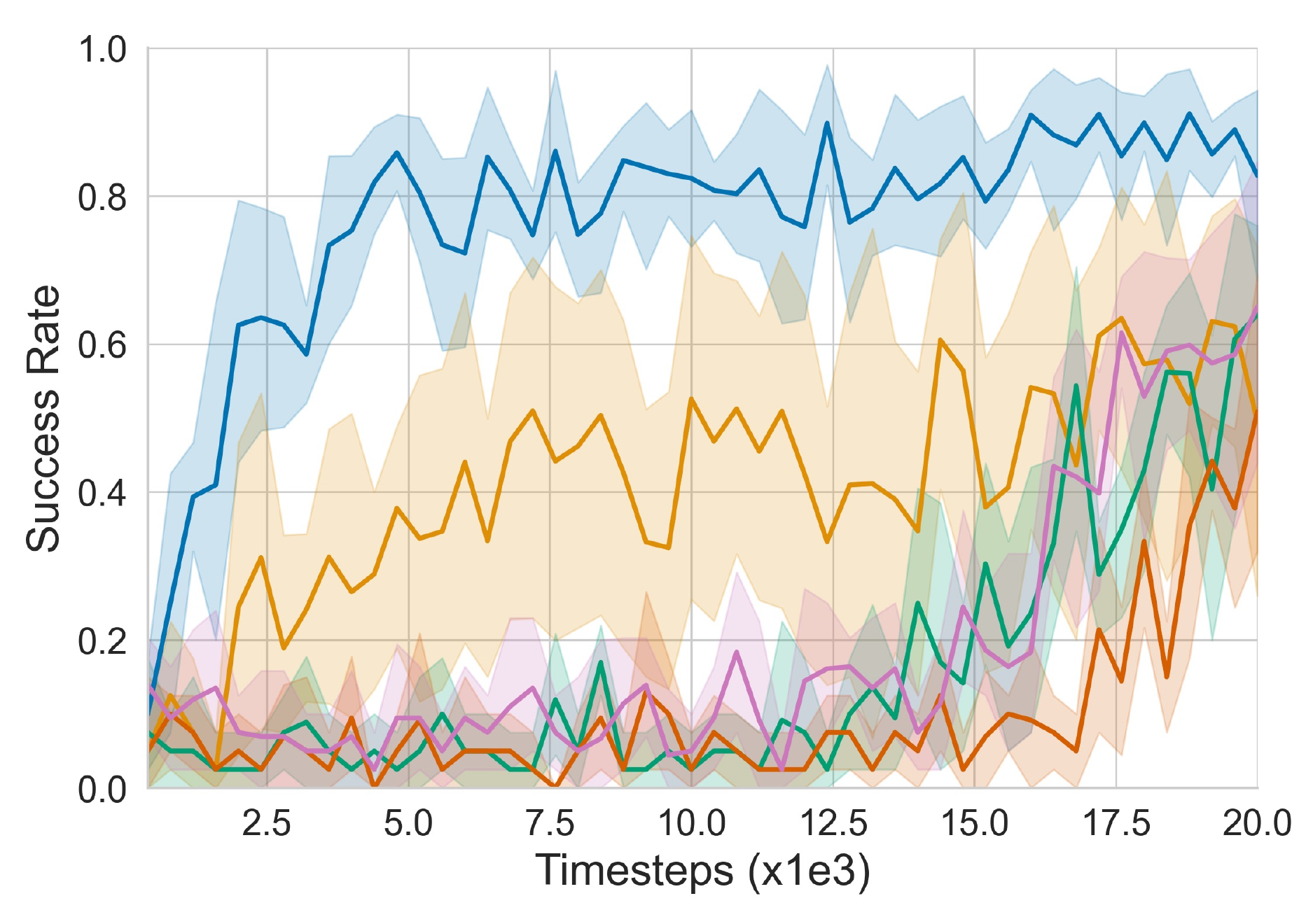} &
			\includegraphics[width=.31\textwidth]{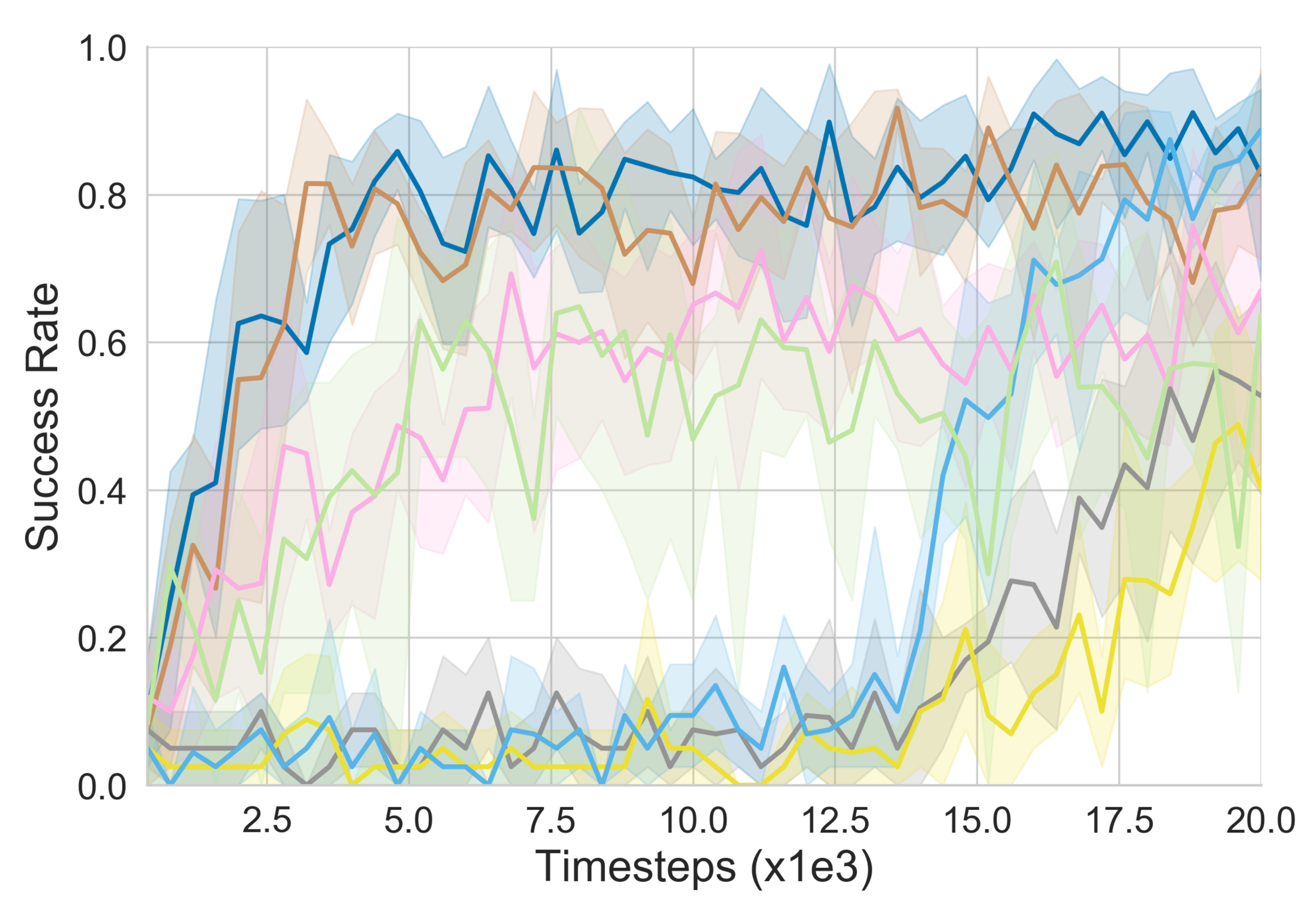} \\
			\small{(a) \texttt{\textbf {$\mathbf 6$D_Reach_5m}}} & \small{(b) Comparison w/ \gls{mm} baselines} & \small{(c) Comparison w/ methods using priors} \\
			
			\includegraphics[width=.30\textwidth]{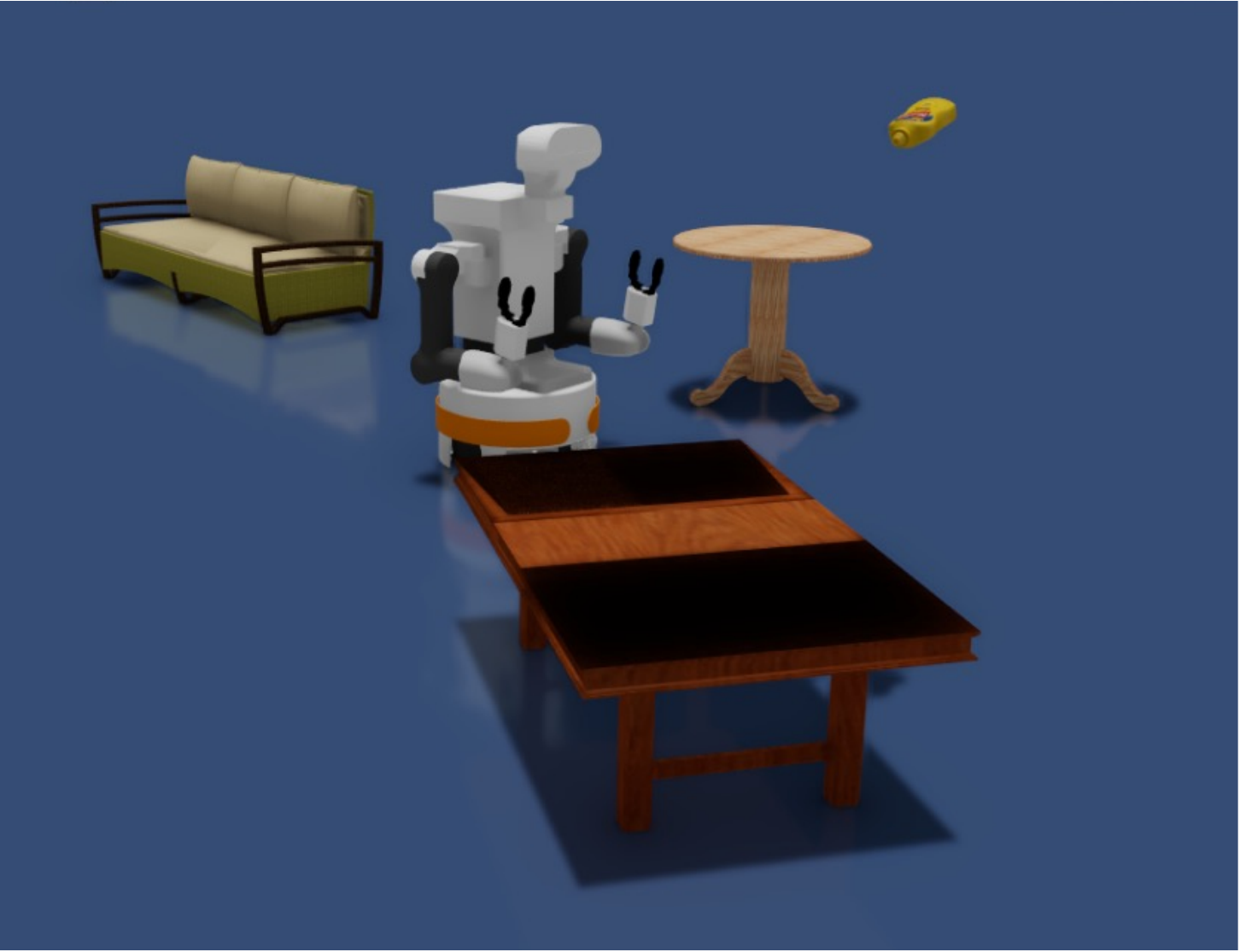} &
			\includegraphics[width=.31\textwidth]{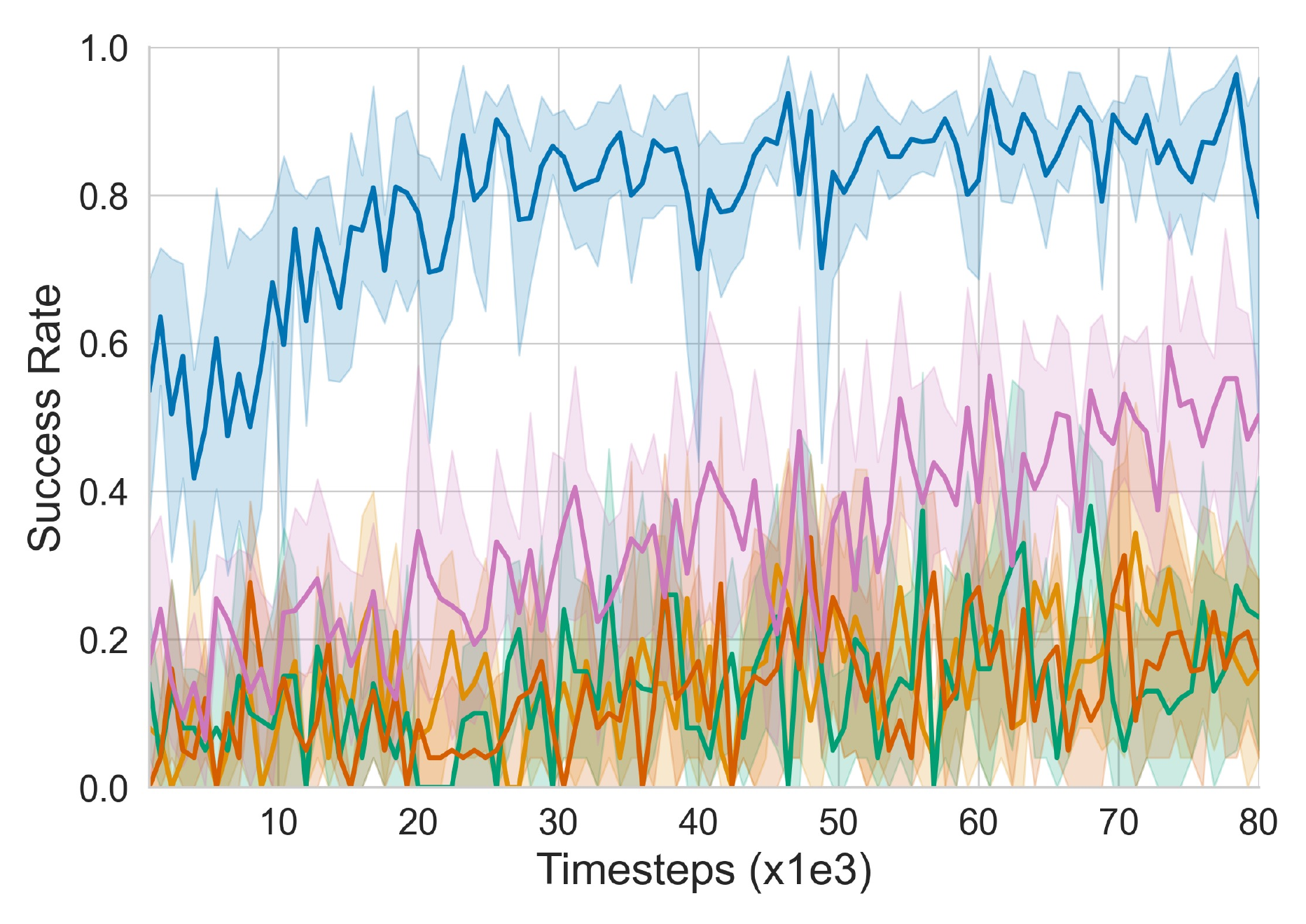} &
			\includegraphics[width=.31\textwidth]{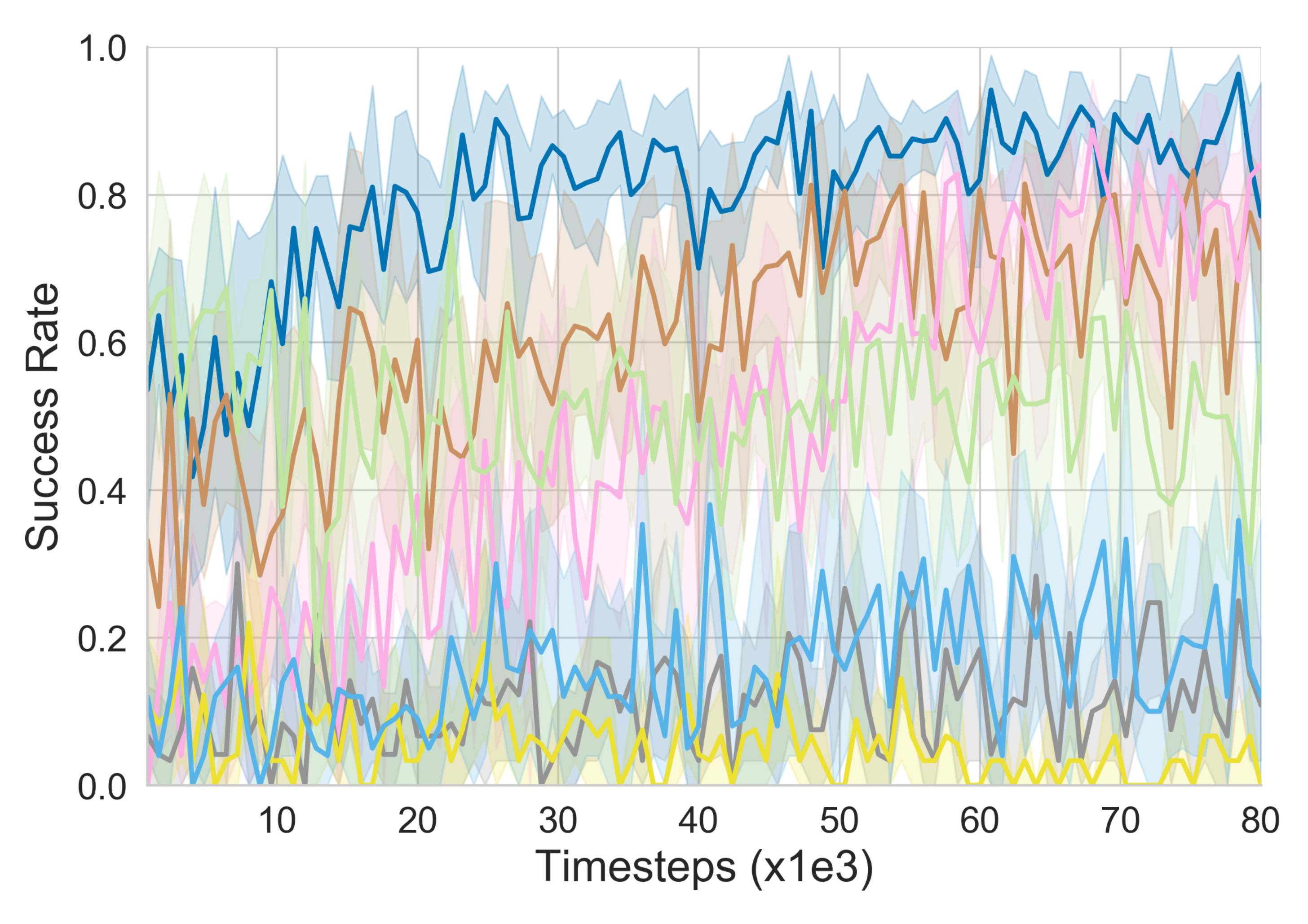} \\
			\small{(d) \texttt{\textbf {$\mathbf 6$D_Reach_3_obstacles}}} & \small{(e) Comparison w/ \gls{mm} baselines} &  \small{(f) Comparison w/ methods using priors} \\
			\multicolumn{3}{c}{\includegraphics[width=\textwidth]{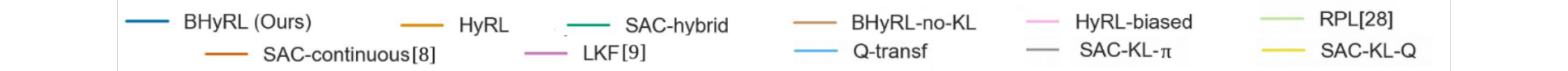}} \\
			
		\end{tabular}
		\caption{\small Snapshots of 6D-reaching environments, and the success rate curves of \gls{bhyrl} and baseline methods, both for learning \gls{mm} and for learning with priors.}
		\label{fig:plots}
		\vspace{+0.35em}
		\captionof{table}{\small Average number of action queries until task completion for the 6D reaching tasks in the $5$m radius and in $4$m radius with $3$ obstacles environments of Fig. \ref{fig:plots}.}
		\adjustbox{max width=\textwidth}{%
			\centering
			\begin{tabular}{*{10}{c|}c}
				\toprule
				\multicolumn{11}{c}{\texttt{\textbf{$\mathbf 6$D_Reach_5m}}: action queries @$20$k steps}\\
				\hline
				\gls{bhyrl} & \gls{hyrl} & \gls{sac}-hybrid & \gls{sac}-continuous & LKF & \gls{bhyrl}-no-\gls{kl} & \gls{hyrl}-biased & \acrshort{rpl} & Q-transf & \gls{sac}-\gls{kl}-$\pi$ & \gls{sac}-\gls{kl}-Q \\ 
				\hline
				$\mathbf{5.70 \pm 0.96}$ & $7.90 \pm 2.00$ & $7.30 \pm 1.82$ & $8.83 \pm 1.21$ & $6.35 \pm 2.3$ & $5.7 \pm 1.00$ & $6.46 \pm 1.44$ & $7.27 \pm 0.1$ & $5.31 \pm 0.85$ & $8.56 \pm 0.72$ & $8.89 \pm 0.91$ \\ \hline \hline
				\multicolumn{11}{c}{\texttt{\textbf{$\mathbf 6$D_Reach_3_obstacles}}: action queries @$80$k steps}\\
				\hline
				$\mathbf{7.38 \pm 1.60}$ & $25.3 \pm 3.01$ & $24.57 \pm 4.22$ & $24.2 \pm 3.01$ & $23.47 \pm 3.98$ & $12.02 \pm 2.76$ & $12.02 \pm 3.01$ & $17.84 \pm 3.52$ & $21.82 \pm 5.96$ & $25.67 \pm 2.84$ & $27.5 \pm 0.0$ \\ 
				\bottomrule
		\end{tabular}}
		\vspace{+0.35em}
		\label{tab: results}
		\vspace{-0.6cm}
	\end{figure*}
	\subsection{Evaluation}
	\subsubsection{MM - Reach} 
	In the reaching tasks, we first compare the learned reachability policy against \gls{irm} as in \cite{vahrenkamp_robot_2013} in a $1$m area that is relative to the operational workspace of TIAGo++. The learned reachability prior is transferred to a $5$m-range reaching task, where the robot has to navigate towards the goal and reach for it. Finally, we devise a more challenging task where the robot has to reach for a 6D target amidst obstacles.
	
	\noindent\texttt{\textbf {$\mathbf 6$D_Reach_1m}}. 
	For this task, we need first to compute the \gls{irm} of TIAGo++. Then, we train a policy and a Q-function, but biasing the data-collection towards high probability reachable poses based on the computed \gls{irm}, and some random actions to avoid overfitting. Fig. \ref{fig:IRMvsHyRL} shows the different maps obtained by the \gls{irm} and the one learned through \gls{hyrl} with \gls{irm} data. As we can observe, the original \gls{irm} struggles to find base poses with high confidence [dark blue]. Our learned Q-function is smoother and more expressive, guiding the agent to areas with high probability of success, hence, yielding superior performance. When comparing \gls{irm} to our learned policy, it obtains a success rate of $73.74$\% against the $100$ \% of the learned one (\gls{hyrl}), over $5$ seeds. Moreover, the greedy querying of \gls{irm} leads to more unnecessary base actions, needing on average $4.8$ action queries to find a good pose, while the learned policy solves the problem sampling only $2.2$ sub-goals. Notably, we trained a \gls{sac} agent with a hybrid action space, without utilizing the data biasing from the \gls{irm}, which also learns a good reachability behavior with a success rate of $91.4$ \%, but requires double the number of samples compared to \gls{hyrl}.
	
	Next, we evaluate our method against representative baselines in the following tasks:
	
	\noindent\texttt{\textbf {$\mathbf 6$D_Reach_5m}}. In this task, we sample environments with a radius of up to $5$m. We transfer the policy and the Q-residual from the previous task \texttt{6D_Reach} to \gls{bhyrl}. The robot has to navigate towards the goal and select the right base pose for activating the arm to reach a random 6D goal. 
	
	\noindent\texttt{\textbf {$\mathbf 6$D_Reach_3obstacles}}. In this task, we simulate three different obstacles, that, at each episode, are randomly placed in a $3$m radius. We transfer the policies learned in \texttt{6D_Reach_5m}. This task shows the effect of the prior policy and Q-residuals amidst \textit{information asymmetry}, i.e., in the new task the state also contains the oriented bounding boxes of obstacles. 
	
	First, we compare \textsl{\gls{bhyrl}} against baselines for learning \gls{mm}, comparing against the following:
	\textit{i.} our \textsl{\gls{hyrl}}, for which we explore by adding Gaussian noise; \textit{ii.} \textsl{\gls{sac}-hybrid}, which is, in essence, the implementation of \gls{hyrl} with maximum entropy exploration; \gls{sac} with no discrete action space (\textsl{\gls{sac}-continuous}), which resembles the method of \cite{xia_relmogen_2021}, where the selection of the embodiment can be tackled by thresholding the continuous value; \textit{iv.} a variant of \gls{lkf} \cite{honerkamp_learning_2021}, in which instead of predicting base velocities, we predict base sub-goals, to be directly comparable to ours, but there is no policy for arm activation -- the IK is queried at every step. Note that we extended all methods to consider 6D goal poses. The learning curves show the superior performance of \gls{bhyrl} against all baseline methods for learning \gls{mm} (Fig. \ref{fig:plots}b \& \ref{fig:plots}e). Notably, from Fig. \ref{fig:plots}b  we can already see that both our \gls{hyrl} and \gls{sac}-hybrid outperform \gls{sac}-continuous (our adapted version of SGP-R \cite{xia_relmogen_2021}), underscoring the benefit of our implementation for hybrid action spaces. Note that for the \texttt{6D_Reach_5m} most methods achieve good performance, but need almost double the amount of samples compared to \gls{bhyrl}, underlying our method's sample-efficiency. In the more challenging task of \texttt{6D_Reach_3_obstacles} (Fig. \ref{fig:plots}e), we observe a notable acceleration in learning compared to the baselines. \gls{bhyrl} allows the agent to learn to avoid obstacles while reaching for the target without the need for precarious exploration. We note the good performance of \gls{lkf} that queries the IK-solver at every step, helping the agent to achieve good success rates, at the cost of more action queries, Table \ref{tab: results}.
	
	Secondly, we consider different ways of incorporating prior information, comparing \textsl{\gls{bhyrl}} against the following algorithms: \textit{i.} \gls{bhyrl} without the \gls{kl} regularizer on the policy (\textsl{\gls{bhyrl}-no-KL}); \textit{ii.} \gls{hyrl} with data collection biasing, where we explore by taking, $50\%$ of the time, actions based on the computed \gls{irm} when we are in the proximity of the goal (\textsl{\gls{hyrl}-biased}); \textit{iii.} \gls{rpl} \cite{silver_residual_2019}, which involves residual learning in policy-space (\textsl{\gls{rpl}}); \textit{iv.} Q-transfer, i.e. a naive transfer of the Q function from a prior task to a SAC agent in a new task (\textsl{Q-transf}). This requires knowing the state-space of downstream tasks a-priori and padding the state accordingly with zeros; \textit{v.} \gls{sac} with a \gls{kl} on the policy (\textsl{\gls{sac}-KL-$\pi$}); \textit{vi.} \gls{sac} with a \gls{kl} on the Q objective (\textsl{\gls{sac}-KL-$Q$});
	
	Our results in Fig. \ref{fig:plots}c \& \ref{fig:plots}f show a clear benefit of Q-residuals against the baselines. The effect of the KL-regularization in combination with the Q-residuals of \gls{bhyrl} is better observed in the more challenging task of Fig. \ref{fig:plots}f. Still, in Fig. \ref{fig:plots}c we can see a slight improvement of \gls{bhyrl} over \gls{bhyrl}-no-\gls{kl}, as the policy regularization provides more expressive behaviors. Crucially, we see that the information asymmetry for transferring from task \texttt{6D_Reach_5m} to \texttt{6D_Reach_3obstacles} is resolved via \gls{bhyrl}. However, both the \gls{kl} regularization of the policy and the \gls{kl} as shaping reward for the Q objective do not benefit \gls{sac} in the challenging task of \texttt{6D_Reach_3obstacles}. The direct transfer of the Q function from the previous task provides a slight acceleration in learning over \gls{sac}. \gls{rpl} performs reasonably well on the \texttt{6D_Reach_5m} task but struggles to improve over the prior policy in the \texttt{6D_Reach_3obstacles} task, highlighting the challenge of learning a residual policy on top of the prior policy's actions. It is noteworthy that in the challenging task with the obstacles, \gls{bhyrl} requires less action queries (sub-goals) to achieve the end-task compared to all other methods, Table \ref{tab: results}.

	\subsubsection{\gls{mm} -- Fetch}
	\begin{figure}[h!]
		\vspace{-0.35em}
		\centering
		\begin{tabular}{c c}
			\includegraphics[width=0.225\textwidth]{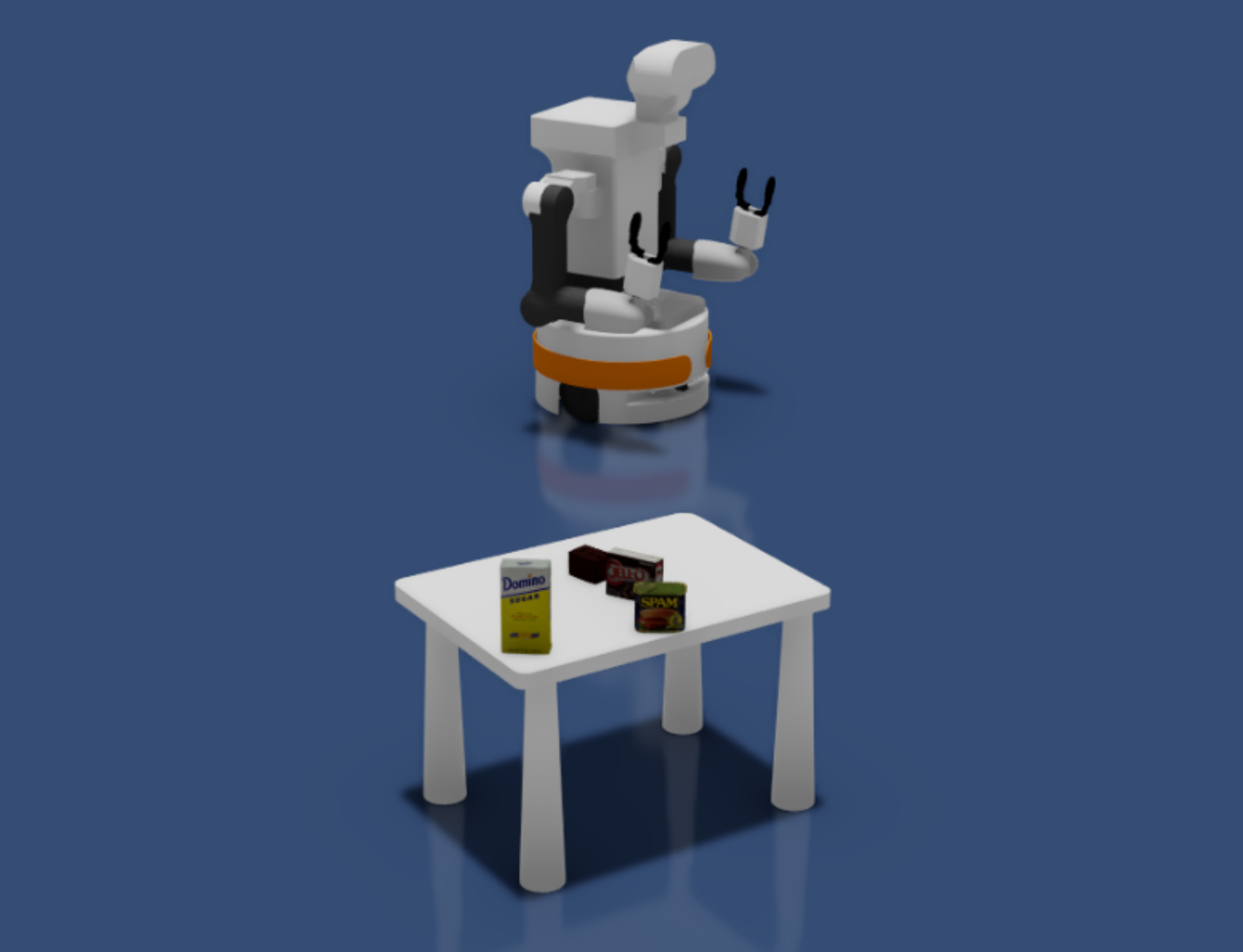}
			&  \includegraphics[width=0.225\textwidth]{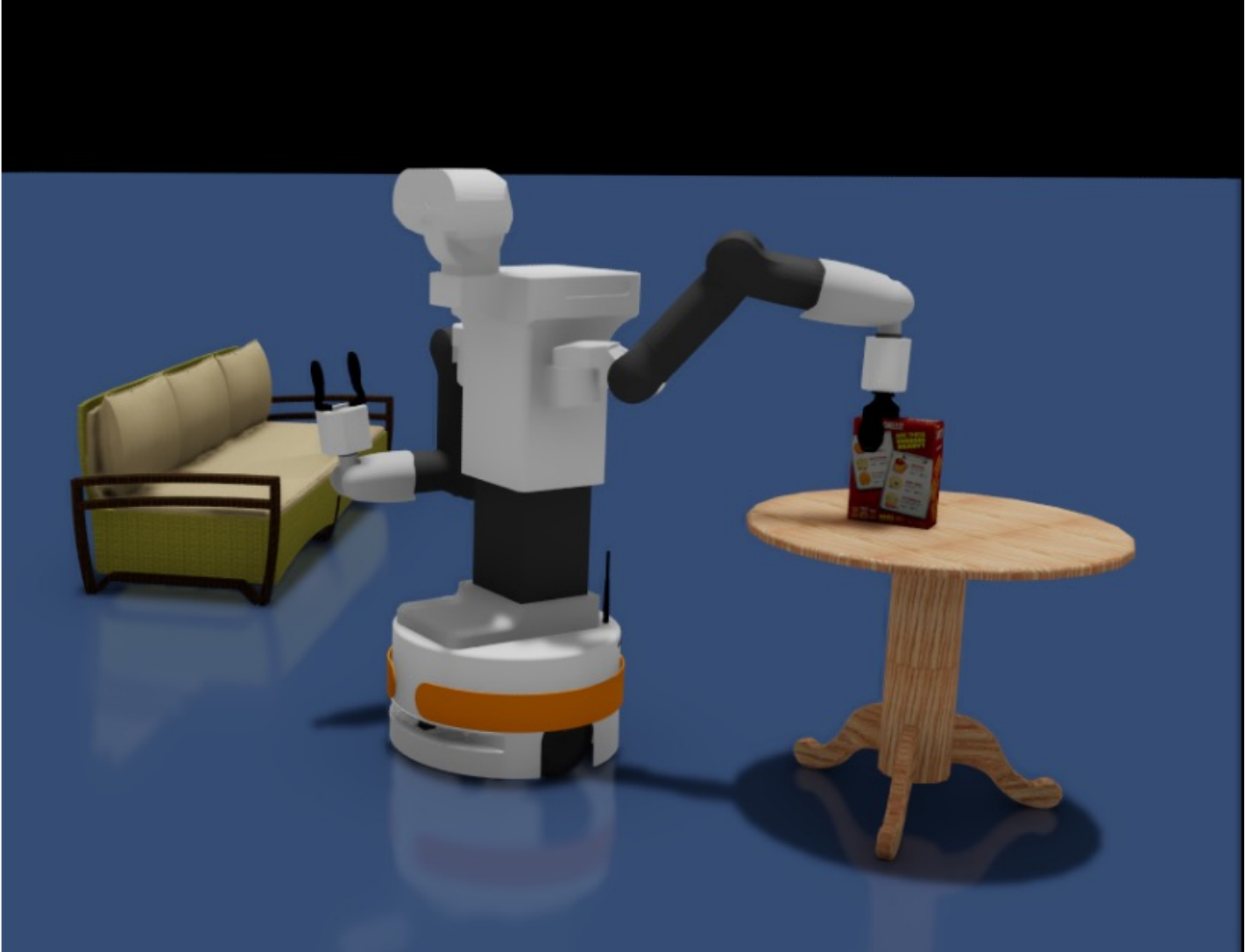} \\
		\end{tabular}
		\captionof{figure}{\texttt{6D_Fetch} environments. \textbf{Left:} 3m-radius environment with multiple grasp objects placed on a table. The relative pose of the table, the objects and their grasps are randomly sampled. \textbf{Right} 4m-radius scene with a table and a sofa that are randomly arranged. Different objects with different grasp poses can appear on the table in random positions at each episode. }
		\label{fig:fetch_task}
		\vspace{-0.4cm}
	\end{figure}
	For this evaluation, we test \gls{bhyrl} on fetching tasks, while required to avoid collisions. We first designed fetching environments in simulation and trained \gls{bhyrl}. Next, we zero-transferred the learned policy to our real TIAGo++ robot for similar fetching and placing tasks.
	
	\noindent\texttt{\textbf {6D_Fetch}}. We designed two final simulated tasks for 6D fetching,  i.e., reaching an object and grasping (Fig. \ref{fig:fetch_task}). For the first task, the goal is a 6D object-grasp to be executed in an environment of $3$m radius. Multiple objects are placed on a table and grasps are randomly sampled from a set of feasible grasps of a target object \cite{eppner2021acronym}. In each episode, we randomly sample a table pose, and the objects are placed randomly over the table. The robot should approach the table without colliding with it, and successfully grasp the target object without colliding with the other objects with its arm. This task shows that the agent can not only learn to avoid collisions with it's base but also learn to place itself with a clearance to avoid arm collisions with other objects. For this task, we transfer the Q-residuals and the policy from \texttt{6D_Reach_5m} to train \gls{bhyrl}. We do not compare to baselines, given their significantly lower performance in the previous tasks. Here, we report an average success rate of $\mathbf{83.27}\%$ at ${200}$k steps for \gls{bhyrl} over $5$ seeds, while the agent completes the task with an average of $\mathbf{5.57}$ action queries. Next, we design a different configuration for our simulated task, with a $4$m-radius scene with a table and a sofa randomly arranged. An object, from a set of three different objects, may appear on top of the table, randomly placed, and a 6D grasp is generated as the goal for the fetching task. At each episode, the scene is randomly arranged. \gls{bhyrl} achieves an average performance of $\mathbf{93.6 \%}$ fetching success rate at ${350}$k steps, and requires on average $\mathbf{8.46}$ action queries (sub-goals) to complete the task.
	
	\begin{figure}[t!]
		\vspace{+0.25em}
		\centering
		\includegraphics[width=\columnwidth]{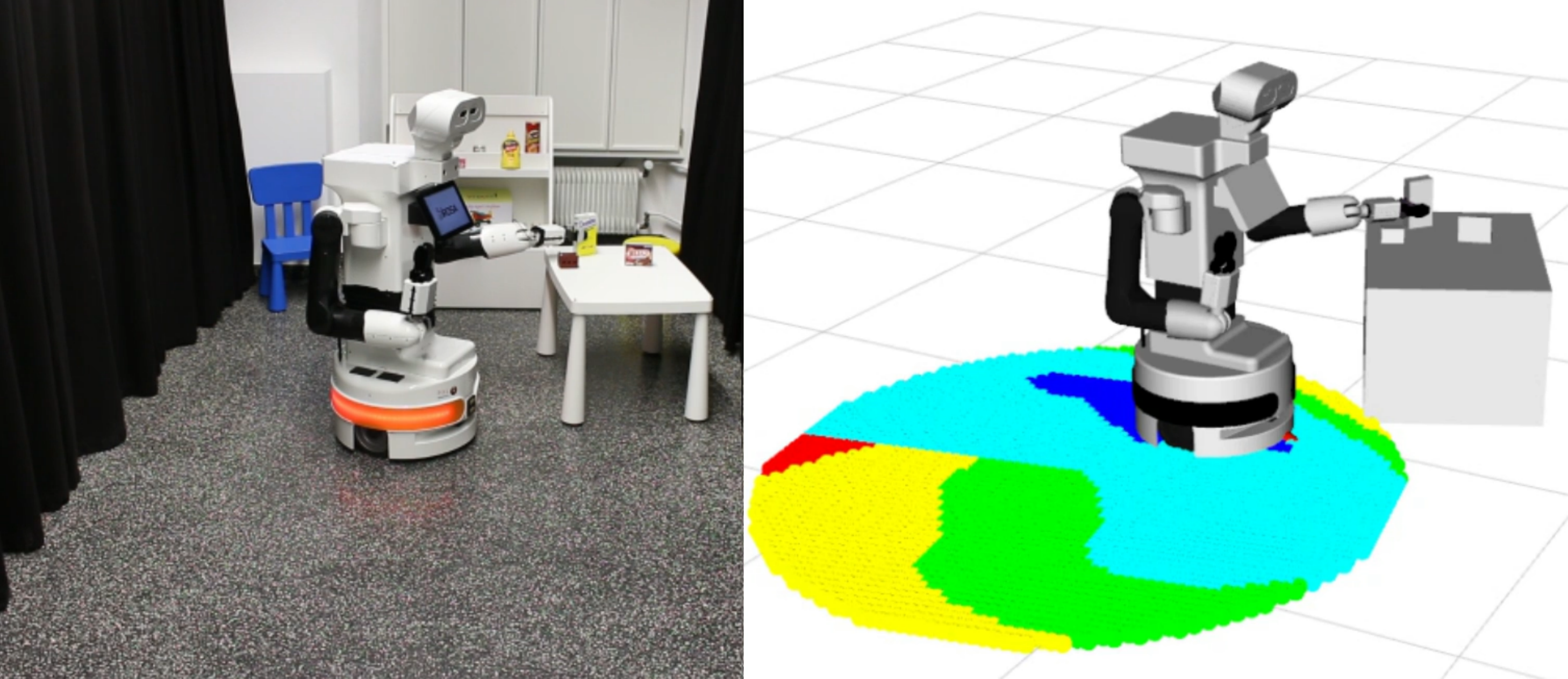}
		\captionof{figure}{Example execution of the real-world 6D fetching task. We perform a zero-shot transfer of the \texttt{6D_Fetch} policy learnt in simulation to the real Tiago++ robot. The visualization on the right shows the learnt base Q function of the robot. Dark {\color{blue}blue} signifies {\color{blue}maximum} likelihood of success while {\color{red}red} signifies the {\color{red}lowest}.}
		\label{fig:pick_tiago}
		\vspace{-0.65cm}
	\end{figure}
	
	\noindent \texttt{\textbf{GoFetch-TIAGo++}}. In this final evaluation, we showcase transferability to our real TIAGo++ robot. Since our state-space includes the relative transform to the robot 6D goal pose and oriented bounding boxes of obstacles, we rely on an OptiTrack motion capture system. Fig. \ref{fig:fetch_tiago} and \ref{fig:pick_tiago} show that the policy can effectively provide sub-goals in the real world, as it is a high-level decision-making process on the robot actions, whose low-level execution relies on well-known motion planning methods for grasp planning and navigating to sub-goals.
	
	We conduct 32 trials with the real-robot system for the fetching task and record a success rate of $\mathbf{81.25}\%$. The robot manages to place itself and successfully plan a motion to grasp the target object querying, on average, $\mathbf{2.2}$ sub-goals. We additionally perform another variant of the experiment: an `object-rearrangement' task. This task involves sequential base-placement for picking up objects from one table and placing them on another flat surface. In Fig. \ref{fig:fetch_tiago}, we depict the learned Q-function for possible poses in \textit{SE}(2), and show how it changes as the robot picks an object from the table and places it on the set of drawers. In almost all trials, all the sub-goals but the last one prevented the robot from using the arm, and only activated it when the robot placed itself in a feasible pose for executing the grasp. Video demonstrations of the robot behavior can be found at: \url{https://irosalab.com/rlmmbp/}.
	
	We observe that the main reason for failure on the real system is the mismatch between the base planner in simulation versus the ROS-based base planner deployed on the real system. In most failed trials, the agent's learned policy output a sub-goal that was close to the table obstacle and the underlying base motion-planner failed to find a safe plan to the sub-goal. This result highlights the importance of not only learning optimal behavior but also considering stronger safety constraints for the real system, which we hope to address in future work.
	
	\section{Conclusions}
	Mobile Manipulation (\gls{mm}) for reaching and fetching tasks requires the effective coordination of the robot's embodiments, as the robot has to choose an optimal base pose and decide on attempting a grasp when fetching an object. 
	In this work, we proposed Boosted Hybrid Reinforcement Learning (\gls{bhyrl}), a novel method for learning \gls{mm} reaching and fetching tasks with reachability behavior priors while considering hybrid action spaces.
	We demonstrated the benefits of our method in simulated tasks of \gls{mm} with increasing difficulty, and we confirmed the superiority of \gls{bhyrl} against baselines both for learning \gls{mm} and for handling priors. Finally, we zero-transferred our 6D fetching policy to our TIAGo++ robot, showing the potential for real-world deployment.
	
	A limitation of this work is that the agent is trained to maximize the likelihood of finding IK solutions, implying optimal actions are learnt only in terms of \textit{reachability} and not \textit{manipulability}, which we wish to address in future work. We also plan to incorporate more visual information and extend \gls{bhyrl} to more challenging tasks such as grasping in clutter and bi-manual \gls{mm} decision-making tasks.
	
	


	
	
	
	\bibliographystyle{IEEEtran}
	\bibliography{svm3,rlwithpriors}

\begin{thebibliography}{10}
\providecommand{\url}[1]{#1}
\csname url@samestyle\endcsname
\providecommand{\newblock}{\relax}
\providecommand{\bibinfo}[2]{#2}
\providecommand{\BIBentrySTDinterwordspacing}{\spaceskip=0pt\relax}
\providecommand{\BIBentryALTinterwordstretchfactor}{4}
\providecommand{\BIBentryALTinterwordspacing}{\spaceskip=\fontdimen2\font plus
\BIBentryALTinterwordstretchfactor\fontdimen3\font minus
  \fontdimen4\font\relax}
\providecommand{\BIBforeignlanguage}[2]{{%
\expandafter\ifx\csname l@#1\endcsname\relax
\typeout{** WARNING: IEEEtran.bst: No hyphenation pattern has been}%
\typeout{** loaded for the language `#1'. Using the pattern for}%
\typeout{** the default language instead.}%
\else
\language=\csname l@#1\endcsname
\fi
#2}}
\providecommand{\BIBdecl}{\relax}
\BIBdecl

\bibitem{brock_mobility_2016}
O.~Brock, J.~Park, and M.~Toussaint, ``Mobility and manipulation,'' in
  \emph{Springer Handbook of Robotics, 2nd Edition}, 2016, pp. 1007--1036.

\bibitem{haarnoja2018soft}
T.~Haarnoja, A.~Zhou, P.~Abbeel, and S.~Levine, ``Soft actor-critic: Off-policy
  maximum entropy deep reinforcement learning with a stochastic actor,'' in
  \emph{{ICML}}, ser. Proceedings of Machine Learning Research, vol.~80, 2018,
  pp. 1856--1865.

\bibitem{ravichandar2020recent}
H.~Ravichandar, A.~Polydoros, S.~Chernova, and A.~Billard, ``Recent advances in
  robot learning from demonstration,'' \emph{Annual Review of Control,
  Robotics, and Autonomous Systems}, vol.~3, pp. 297--330, 2020.

\bibitem{ibarz2021train}
J.~Ibarz, J.~Tan, C.~Finn, M.~Kalakrishnan, P.~Pastor, and S.~Levine, ``How to
  train your robot with deep reinforcement learning: lessons we have learned,''
  \emph{IJRR}, vol.~40, no. 4-5, pp. 698--721, 2021.

\bibitem{lee_learning_2020}
J.~Lee, J.~Hwangbo, L.~Wellhausen, V.~Koltun, and M.~Hutter, ``Learning
  quadrupedal locomotion over challenging terrain,'' \emph{Sci. Robotics},
  vol.~5, no.~47, p. 5986, 2020.

\bibitem{dalal_accelerating_2021}
M.~Dalal, D.~Pathak, and R.~R. Salakhutdinov, ``Accelerating robotic
  reinforcement learning via parameterized action primitives,'' \emph{Advances
  in Neural Information Processing Systems}, vol.~34, pp. 21\,847--21\,859,
  2021.

\bibitem{li_hrl4in_2019}
C.~Li, F.~Xia, R.~Mart{\'{\i}}n{-}Mart{\'{\i}}n, and S.~Savarese, ``{HRL4IN:}
  hierarchical reinforcement learning for interactive navigation with mobile
  manipulators,'' in \emph{CoRL}, ser. PMLR, vol. 100, 2019, pp. 603--616.

\bibitem{xia_relmogen_2021}
F.~Xia, C.~Li, R.~Mart{\'{\i}}n{-}Mart{\'{\i}}n, O.~Litany, A.~Toshev, and
  S.~Savarese, ``Relmogen: Integrating motion generation in reinforcement
  learning for mobile manipulation,'' in \emph{{ICRA}}.\hskip 1em plus 0.5em
  minus 0.4em\relax {IEEE}, 2021, pp. 4583--4590.

\bibitem{honerkamp_learning_2021}
D.~Honerkamp, T.~Welschehold, and A.~Valada, ``Learning kinematic feasibility
  for mobile manipulation through deep reinforcement learning,'' \emph{{IEEE}
  Robotics Autom. Lett.}, vol.~6, no.~4, pp. 6289--6296, 2021.

\bibitem{nair2020awac}
A.~Nair, M.~Dalal, A.~Gupta, and S.~Levine, ``Accelerating online reinforcement
  learning with offline datasets,'' \emph{CoRR}, vol. abs/2006.09359, 2020.

\bibitem{johannink2019residual}
T.~Johannink, S.~Bahl, A.~Nair, J.~Luo, A.~Kumar, M.~Loskyll, J.~A. Ojea,
  E.~Solowjow, and S.~Levine, ``Residual reinforcement learning for robot
  control,'' in \emph{{ICRA}}.\hskip 1em plus 0.5em minus 0.4em\relax {IEEE},
  2019, pp. 6023--6029.

\bibitem{morgan2021model}
A.~S. Morgan, D.~Nandha, G.~Chalvatzaki, C.~D'Eramo, A.~M. Dollar, and
  J.~Peters, ``Model predictive actor-critic: Accelerating robot skill
  acquisition with deep reinforcement learning,'' in \emph{{ICRA}}.\hskip 1em
  plus 0.5em minus 0.4em\relax {IEEE}, 2021, pp. 6672--6678.

\bibitem{wolfe_combined_2010}
J.~A. Wolfe, B.~Marthi, and S.~Russell, ``Combined task and motion planning for
  mobile manipulation,'' in \emph{{ICAPS}}.\hskip 1em plus 0.5em minus
  0.4em\relax {AAAI}, 2010, pp. 254--258.

\bibitem{vahrenkamp_robot_2013}
N.~Vahrenkamp, T.~Asfour, and R.~Dillmann, ``Robot placement based on
  reachability inversion,'' in \emph{{ICRA}}.\hskip 1em plus 0.5em minus
  0.4em\relax {IEEE}, 2013, pp. 1970--1975.

\bibitem{makhal_reuleaux_2018}
A.~Makhal and A.~K. Goins, ``Reuleaux: Robot base placement by reachability
  analysis,'' in \emph{{IRC}}.\hskip 1em plus 0.5em minus 0.4em\relax {IEEE}
  Computer Society, 2018, pp. 137--142.

\bibitem{sun2022fully}
C.~Sun, J.~Orbik, C.~M. Devin, B.~H. Yang, A.~Gupta, G.~Berseth, and S.~Levine,
  ``Fully autonomous real-world reinforcement learning with applications to
  mobile manipulation,'' in \emph{CoRL}, ser. Proceedings of Machine Learning
  Research, vol. 164.\hskip 1em plus 0.5em minus 0.4em\relax {PMLR}, 2021, pp.
  308--319.

\bibitem{galashov2019information}
A.~Galashov, S.~M. Jayakumar, L.~Hasenclever, D.~Tirumala, J.~Schwarz,
  G.~Desjardins, W.~M. Czarnecki, Y.~W. Teh, R.~Pascanu, and N.~Heess,
  ``Information asymmetry in kl-regularized {RL},'' \emph{CoRR}, vol.
  abs/1905.01240, 2019.

\bibitem{roa_mobile_2021}
M.~A. Roa, M.~R. Dogar, J.~Pag{\`{e}}s, C.~Vivas, A.~Morales, N.~Correll,
  M.~Gorner, J.~Rosell, S.~Foix, R.~Memmesheimer, and F.~Ferro, ``Mobile
  manipulation hackathon: Moving into real world applications,'' \emph{{IEEE}
  Robotics Autom. Mag.}, vol.~28, no.~2, pp. 112--124, 2021.

\bibitem{vahrenkamp_representing_2015}
N.~Vahrenkamp and T.~Asfour, ``Representing the robot's workspace through
  constrained manipulability analysis,'' \emph{Auton. Robots}, vol.~38, no.~1,
  pp. 17--30, 2015.

\bibitem{hertle_identifying_2017}
A.~Hertle and B.~Nebel, ``Identifying good poses when doing your household
  chores: Creation and exploitation of inverse surface reachability maps,'' in
  \emph{{IROS}}.\hskip 1em plus 0.5em minus 0.4em\relax {IEEE}, 2017, pp.
  6053--6058.

\bibitem{welschehold_coupling_2018}
T.~Welschehold, C.~Dornhege, F.~Paus, T.~Asfour, and W.~Burgard, ``Coupling
  mobile base and end-effector motion in task space,'' in \emph{{IROS}}.\hskip
  1em plus 0.5em minus 0.4em\relax {IEEE}, 2018, pp. 1--9.

\bibitem{chitta_mobile_2012}
S.~Chitta, E.~G. Jones, M.~T. Ciocarlie, and K.~Hsiao, ``Mobile manipulation in
  unstructured environments: Perception, planning, and execution,''
  \emph{{IEEE} Robotics Autom. Mag.}, vol.~19, no.~2, pp. 58--71, 2012.

\bibitem{burget_bi_2016}
F.~Burget, M.~Bennewitz, and W.~Burgard, ``Bi\({}^{\mbox{2}}\)rrt*: An
  efficient sampling-based path planning framework for task-constrained mobile
  manipulation,'' in \emph{{IROS}}.\hskip 1em plus 0.5em minus 0.4em\relax
  {IEEE}, 2016, pp. 3714--3721.

\bibitem{welschehold_learning_2017}
T.~Welschehold, C.~Dornhege, and W.~Burgard, ``Learning mobile manipulation
  actions from human demonstrations,'' in \emph{{IROS}}.\hskip 1em plus 0.5em
  minus 0.4em\relax {IEEE}, 2017, pp. 3196--3201.

\bibitem{welschehold_combined_2019}
T.~Welschehold, N.~Abdo, C.~Dornhege, and W.~Burgard, ``Combined task and
  action learning from human demonstrations for mobile manipulation
  applications,'' in \emph{{IROS}}.\hskip 1em plus 0.5em minus 0.4em\relax
  {IEEE}, 2019, pp. 4317--4324.

\bibitem{kindle_whole-body_2020}
J.~Kindle, F.~Furrer, T.~Novkovic, J.~J. Chung, R.~Siegwart, and J.~I. Nieto,
  ``Whole-body control of a mobile manipulator using end-to-end reinforcement
  learning,'' \emph{CoRR}, vol. abs/2003.02637, 2020.

\bibitem{mittal2021articulated}
M.~Mittal, D.~Hoeller, F.~Farshidian, M.~Hutter, and A.~Garg, ``Articulated
  object interaction in unknown scenes with whole-body mobile manipulation,''
  \emph{CoRR}, vol. abs/2103.10534, 2021.

\bibitem{silver_residual_2019}
T.~Silver, K.~R. Allen, J.~Tenenbaum, and L.~P. Kaelbling, ``Residual policy
  learning,'' \emph{CoRR}, vol. abs/1812.06298, 2018.

\bibitem{rana_bayesian_2021}
K.~Rana, V.~Dasagi, J.~Haviland, B.~Talbot, M.~Milford, and
  N.~S{\"{u}}nderhauf, ``Bayesian controller fusion: Leveraging control priors
  in deep reinforcement learning for robotics,'' \emph{CoRR}, vol.
  abs/2107.09822, 2021.

\bibitem{rana_multiplicative_2020}
K.~Rana, V.~Dasagi, B.~Talbot, M.~Milford, and N.~S{\"{u}}nderhauf,
  ``Multiplicative controller fusion: Leveraging algorithmic priors for
  sample-efficient reinforcement learning and safe sim-to-real transfer,'' in
  \emph{{IROS}}.\hskip 1em plus 0.5em minus 0.4em\relax {IEEE}, 2020, pp.
  6069--6076.

\bibitem{peters2010relative}
J.~Peters, K.~M{\"{u}}lling, and Y.~Altun, ``Relative entropy policy search,''
  in \emph{{AAAI}}.\hskip 1em plus 0.5em minus 0.4em\relax {AAAI} Press, 2010.

\bibitem{abdolmaleki2018maximum}
A.~Abdolmaleki, J.~T. Springenberg, Y.~Tassa, R.~Munos, N.~Heess, and M.~A.
  Riedmiller, ``Maximum a posteriori policy optimisation,'' \emph{CoRR}, vol.
  abs/1806.06920, 2018.

\bibitem{cheng2019control}
R.~Cheng, A.~Verma, G.~Orosz, S.~Chaudhuri, Y.~Yue, and J.~Burdick, ``Control
  regularization for reduced variance reinforcement learning,'' in
  \emph{{ICML}}, ser. Proceedings of Machine Learning Research, vol.~97.\hskip
  1em plus 0.5em minus 0.4em\relax {PMLR}, 2019, pp. 1141--1150.

\bibitem{jeong2020learning}
R.~Jeong, J.~T. Springenberg, J.~Kay, D.~Zheng, A.~Galashov, N.~Heess, and
  F.~Nori, ``Learning dexterous manipulation from suboptimal experts,'' in
  \emph{CoRL}, ser. Proceedings of Machine Learning Research, vol. 155.\hskip
  1em plus 0.5em minus 0.4em\relax {PMLR}, 2020, pp. 915--934.

\bibitem{pertsch_accelerating_2020}
K.~Pertsch, Y.~Lee, and J.~J. Lim, ``Accelerating reinforcement learning with
  learned skill priors,'' in \emph{CoRL}, ser. Proceedings of Machine Learning
  Research, vol. 155.\hskip 1em plus 0.5em minus 0.4em\relax {PMLR}, 2020, pp.
  188--204.

\bibitem{neunert2020continuous}
M.~Neunert, A.~Abdolmaleki, M.~Wulfmeier, T.~Lampe, J.~T. Springenberg,
  R.~Hafner, F.~Romano, J.~Buchli, N.~Heess, and M.~A. Riedmiller,
  ``Continuous-discrete reinforcement learning for hybrid control in
  robotics,'' in \emph{CoRL}, ser. Proceedings of Machine Learning Research,
  vol. 100.\hskip 1em plus 0.5em minus 0.4em\relax {PMLR}, 2019, pp. 735--751.

\bibitem{maddison2016concrete}
C.~J. Maddison, A.~Mnih, and Y.~W. Teh, ``The concrete distribution: {A}
  continuous relaxation of discrete random variables,'' \emph{CoRR}, vol.
  abs/1611.00712, 2016.

\bibitem{jang2016categorical}
E.~Jang, S.~Gu, and B.~Poole, ``Categorical reparameterization with
  gumbel-softmax,'' \emph{CoRR}, vol. abs/1611.01144, 2016.

\bibitem{klink2022boosted}
P.~Klink, C.~D'Eramo, J.~Peters, and J.~Pajarinen, ``Boosted curriculum
  reinforcement learning,'' in \emph{International Conference on Learning
  Representations}, 2022.

\bibitem{freund1995boosting}
Y.~Freund, ``Boosting a weak learning algorithm by majority,'' \emph{Inf.
  Comput.}, vol. 121, no.~2, pp. 256--285, 1995.

\bibitem{tosatto2017boosted}
S.~Tosatto, M.~Pirotta, C.~D'Eramo, and M.~Restelli, ``Boosted fitted
  q-iteration,'' in \emph{{ICML}}, ser. Proceedings of Machine Learning
  Research, vol.~70.\hskip 1em plus 0.5em minus 0.4em\relax {PMLR}, 2017, pp.
  3434--3443.

\bibitem{MushroomJMLR}
C.~D'Eramo, D.~Tateo, A.~Bonarini, M.~Restelli, and J.~Peters, ``Mushroomrl:
  Simplifying reinforcement learning research,'' \emph{JMLR}, vol.~22, pp.
  131:1--131:5, 2021.

\bibitem{eppner2021acronym}
C.~Eppner, A.~Mousavian, and D.~Fox, ``{ACRONYM:} {A} large-scale grasp dataset
  based on simulation,'' in \emph{{ICRA}}.\hskip 1em plus 0.5em minus
  0.4em\relax {IEEE}, 2021, pp. 6222--6227.

\end{thebibliography}

\end{document}